\documentclass[letterpaper,10pt,journal,twoside]{IEEEtran}

\usepackage{graphicx}
\usepackage{amsmath}
\usepackage{amssymb}
\usepackage{cite}
\usepackage{flafter}
\usepackage{booktabs}
\usepackage{multirow}
\makeatletter
\let\NAT@parse\undefined
\makeatother
\usepackage[colorlinks=true,allcolors=blue,urlcolor=black]{hyperref}
\usepackage{orcidlink}

\title{
NIF-VIO: Visual--Inertial Odometry Across Inertial and
Non-Inertial Reference Frames
}

\author{Zichao Shu$^{1}$~\orcidlink{0009-0000-4785-6234},
Lijun Li$^{2}$~\orcidlink{0009-0007-0818-6929},
Zhuochen Li$^{3}$,
Zetao Chen$^{4}$~\orcidlink{0000-0002-5596-5008},
and Jianyu Wang$^{1}$%
\thanks{\textit{Corresponding author: Zetao Chen.}}%
\thanks{$^{1}$Zichao Shu and Jianyu Wang are with the Shanghai Institute of Technical Physics, Chinese Academy of Sciences, Shanghai 200083, China, and also with the University of Chinese Academy of Sciences, Beijing 100049, China (e-mail: \href{mailto:shuzichao25@mails.ucas.ac.cn}{shuzichao25@mails.ucas.ac.cn}; \href{mailto:jywang@mail.sitp.ac.cn}{jywang@mail.sitp.ac.cn}).}%
\thanks{$^{2}$Lijun Li is with GravityXR, Ningbo 315040, China (e-mail: \href{mailto:ljli@gravityxr.cn}{ljli@gravityxr.cn}).}%
\thanks{$^{3}$Zhuochen Li is with the Advanced Display and Sensing Research Center, Yongjiang Laboratory, Ningbo 315201, China, and also with the University of Nottingham Ningbo China, Ningbo 315100, China (e-mail: \href{mailto:Zhuochen.LI@nottingham.edu.cn}{Zhuochen.LI@nottingham.edu.cn}).}%
\thanks{$^{4}$Zetao Chen is with the Advanced Display and Sensing Research Center, Yongjiang Laboratory, Ningbo 315201, China (e-mail: \href{mailto:zetao-chen@ylab.ac.cn}{zetao-chen@ylab.ac.cn}).}%
}

\hypersetup{pdftitle={NIF-VIO: Visual--Inertial Odometry Across Inertial and Non-Inertial Reference Frames},pdfauthor={Zichao Shu, Lijun Li, Zhuochen Li, Zetao Chen, Jianyu Wang}}

\begin{document}

\bstctlcite{NIFVIO:BSTcontrol}

\maketitle

\begin{abstract}

Conventional visual--inertial odometry (VIO) assumes that observations from cameras and an inertial measurement unit (IMU) are jointly consistent in an inertial reference frame. However, this assumption breaks down under non-inertial conditions. Inside accelerating or rotating carriers, visual observations of the interior reflect carrier-relative motion, while IMU measurements contain additional carrier-induced components, which can compromise conventional fusion. We therefore present NIF-VIO for localization relative to the local environment across inertial and non-inertial reference frames. It adapts visual--inertial fusion based on measurement consistency. As inertial conditions change, early constraint isolation is combined with local reoptimization to limit error propagation before inconsistency confirmation and repair the affected local geometry. Under non-inertial conditions, a compact model with temporally shared parameters is introduced to account for carrier motion while preserving inertial information that complements visual constraints. Experiments on public benchmarks and a self-collected dataset demonstrate that NIF-VIO achieves competitive performance under inertial conditions and improved localization accuracy and availability under non-inertial conditions. The code and dataset will be publicly available upon acceptance.

\end{abstract}


\section{INTRODUCTION}
\label{sec:introduction}

\begin{figure}[t]
\centering
\includegraphics[width=0.96\linewidth]{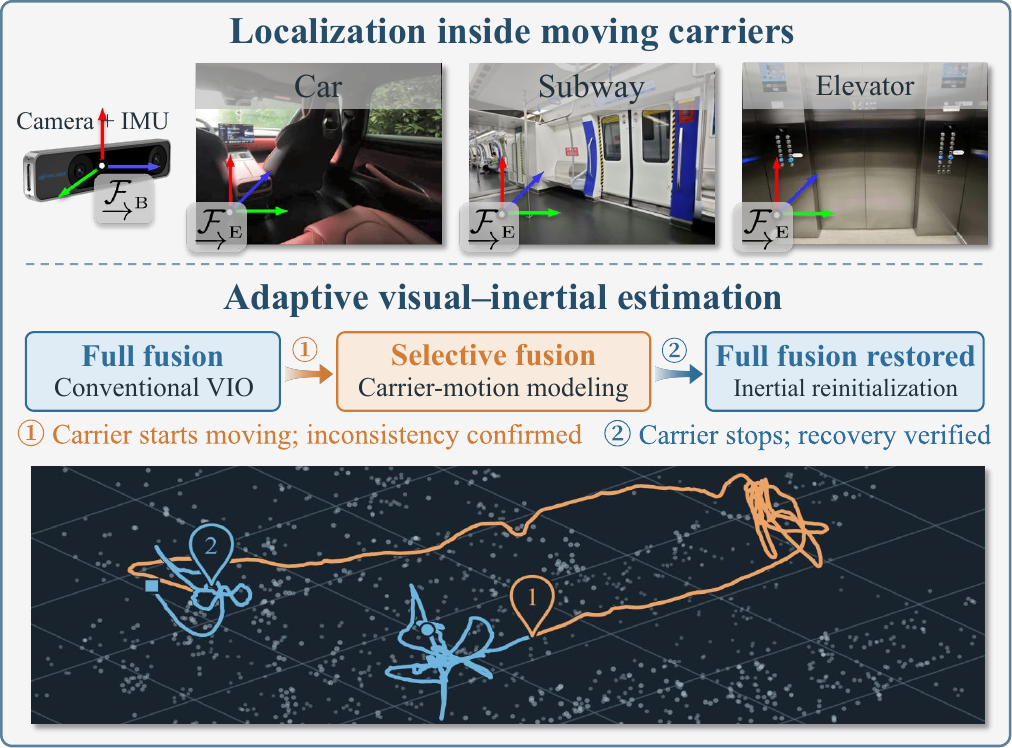}
\caption{Conceptual overview of NIF-VIO. The upper panel shows typical carrier interiors where non-inertial conditions can arise, while the lower panel illustrates adaptive visual--inertial fusion for continuous motion estimation relative to the local environment. Blue and orange denote stationary and moving carrier phases, respectively.}
\label{fig:teaser}
\end{figure}

\IEEEPARstart{V}{isual}--inertial odometry (VIO) combines complementary visual and inertial measurements to estimate ego-motion~\cite{Qin2018VINSMono,Campos2021ORBSLAM3,Stumberg2022DMVIO,Peng2025SqrtVINS}. Conventional VIO assumes that the observed environment is stationary in an inertial reference frame, enabling consistent fusion of the two modalities~\cite{Song2024DynaVINSpp}. However, real-world operation may extend to settings inside moving carriers such as cars, trains, and elevators~\cite{Cortes2018ADVIO,Krishnan2025LaMARiA,Chang2026NoninertialSurvey}. For localization tasks defined relative to the local environment, the reference frame is fixed to the carrier and becomes non-inertial when the carrier accelerates or rotates. Visual observations of the interior provide information about device motion relative to the carrier, whereas measurements from an inertial measurement unit (IMU) also include carrier-induced effects. This mismatch leads to inconsistent constraints under the conventional VIO model and can cause severe estimation errors or even failure~\cite{Lai2024DualIMU,Krishnan2025LaMARiA}.

Visual odometry (VO) can avoid this conflict by relying on vision alone~\cite{Judd2024MultimotionVO}. However, IMU measurements can improve robustness under inertial conditions by complementing visual observations over short time scales~\cite{Boche2025OKVIS2X}. With appropriate modeling, they can also contribute to relative state estimation under non-inertial conditions~\cite{Mandali2026ProprioceptiveNoninertial}. For carrier-relative localization, some existing methods use an additional carrier-mounted IMU to account for carrier motion and retain inertial constraints, but require carrier-side instrumentation and data access~\cite{Lai2024DualIMU,Xia2025DualPreintegration}. Observation selection offers another way to handle visual--inertial conflicts~\cite{Babu2018MotionConflict}. A multi-motion extension supports carrier-relative tracking, but the corresponding pose optimization still relies on visual constraints alone~\cite{Babu2018PerPixelMotion}. These considerations motivate a visual--inertial framework that uses only device-side sensors to maintain continuous localization while retaining useful inertial constraints across changing inertial conditions.

To this end, we propose NIF-VIO. Fig.~\ref{fig:teaser} illustrates its target scenarios and how it adapts fusion as inertial conditions change. Our main contributions are as follows:
\begin{itemize}
\item We propose a unified VIO framework for continuous localization relative to the local environment across inertial and non-inertial conditions, without carrier-side sensing.
\item We develop a transition-handling strategy that combines early constraint isolation with local geometric repair to mitigate error propagation and geometric distortion caused by delayed inconsistency confirmation.
\item We formulate selective inertial constraints through a structured carrier-motion model with temporally shared parameters and soft motion priors, preserving useful inertial information for relative localization under non-inertial conditions.
\item We introduce a dataset spanning car, subway, and elevator scenarios, with carrier-relative reference trajectories for benchmarking localization under changing inertial conditions.
\end{itemize}

\section{RELATED WORK}
\label{sec:related_work}

This section first reviews conventional visual and visual--inertial odometry, then discusses localization and benchmarking in non-inertial environments.

\subsection{Visual and Visual--Inertial Odometry}

Existing VIO methods mainly follow two paradigms: filtering and nonlinear optimization. Filtering-based methods recursively estimate the state through propagation and measurement updates~\cite{Geneva2020OpenVINS,Fornasier2024MSCEqF,Peng2025SqrtVINS}, while optimization-based methods jointly refine states by minimizing visual and inertial residuals, typically over a sliding window~\cite{Qin2018VINSMono,Usenko2020Basalt,Stumberg2022DMVIO,Jung2026ASPIREVINS}. Despite these differences, both paradigms generally rely on visual and inertial measurements being consistent with device motion in an inertial reference frame.

When observations become inconsistent, existing methods mitigate their effects through robust residual treatment, consistency-based fusion selection, and state recovery~\cite{Han2025POAGNC,Lanegger2024ToFuse,Song2024DynaVINSpp}. In non-inertial environments, however, inconsistency can arise from the different motions represented by visual and inertial measurements rather than unreliable observations. When localization is defined relative to a moving carrier, visual observations describe device motion relative to its interior, whereas IMU measurements also contain carrier-induced components. Visual-only methods avoid this conflict and support carrier-relative estimation given sufficient visual constraints from the rigid interior~\cite{Campos2021ORBSLAM3,Judd2024MultimotionVO,Qin2019VINSFusion}, but lose the benefits of IMU measurements. This motivates adapting the use of inertial information to operating conditions rather than discarding it entirely.

\subsection{Localization in Non-Inertial Environments}

In non-inertial environments, one line of work uses an additional carrier-side IMU to account for carrier motion in relative localization, but requires carrier instrumentation and data access~\cite{Lai2024DualIMU,Xia2025DualPreintegration}. Without carrier-side measurements, MC-VIO detects visual--inertial motion conflicts and selects compatible observations for inertial-frame localization~\cite{Babu2018MotionConflict}. Its multi-motion extension supports carrier-relative tracking, but requires user initiation and estimates the corresponding pose from visual constraints alone~\cite{Babu2018PerPixelMotion}. Other work switches between visual and visual--inertial estimation based on measurement consistency, reducing inertial weights before entering visual-only mode and briefly running both estimators in parallel to verify sustained agreement before restoring fusion~\cite{Dine2024InertialDataManagement}. Alternatively, a joint estimation approach augments the device state with carrier-motion states and estimates them through recursive prediction and update using measurements from cameras and an IMU on the device~\cite{Elsdon2025MovingPlatform}. For elevator scenarios, Elevator-LIO and Elevator-VIGS model vertical motion to address carrier-induced conflicts and recover trajectories across floors in a fixed world frame~\cite{Zhang2026ElevatorLIO,Zhou2026ElevatorVIGS}.

Beyond algorithm design, evaluating carrier-relative localization requires datasets with reference trajectories expressed in the carrier frame. ADVIO and UMA-VI include elevator sequences involving non-inertial motion, but the former provides building-frame trajectories and the latter offers reference poses mainly near sequence endpoints~\cite{Cortes2018ADVIO,ZunigaNoel2020UMAVI}. LaMAria includes moving-platform sequences and evaluates trajectories using control points in a fixed world frame. However, the accuracy of its dense pseudo-ground-truth trajectories is not guaranteed for these sequences~\cite{Krishnan2025LaMARiA}. The dataset accompanying Elevator-LIO likewise targets world-frame trajectories across floors rather than motion relative to the elevator itself~\cite{Zhang2026ElevatorLIO}. These differences limit their direct use for evaluating carrier-relative localization. To complement these datasets, we provide a dataset covering car, subway, and elevator scenarios, with carrier-relative reference trajectories spanning changes in inertial conditions.

\section{METHODOLOGY}
\label{sec:methodology}

\begin{figure*}[t]
\centering
\includegraphics[width=0.9\textwidth]{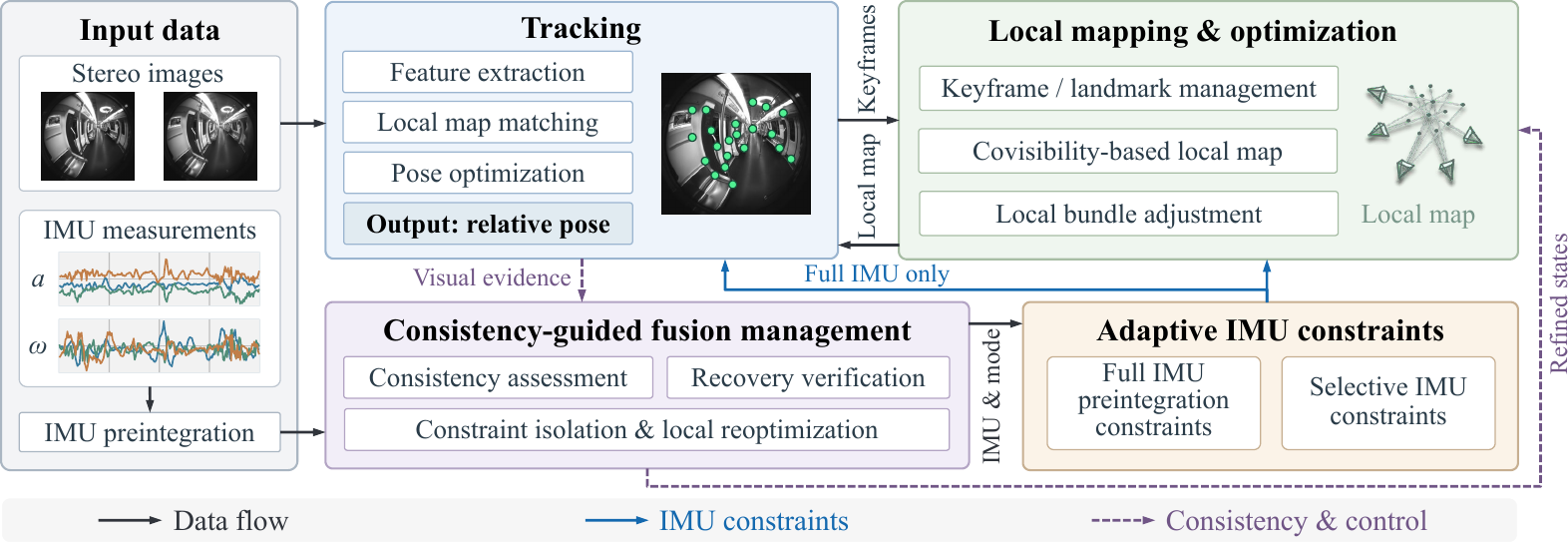}
\caption{System overview of NIF-VIO. Building on a tracking and local mapping architecture, the framework uses visual--inertial consistency to guide IMU fusion. Transition management supports reliable estimation as inertial conditions change, while selective inertial constraints preserve useful IMU information for relative-motion estimation under non-inertial conditions.}
\label{fig:system_overview}
\end{figure*}

Fig.~\ref{fig:system_overview} illustrates the processing pipeline of NIF-VIO. The tracking and local mapping modules follow a structure similar to that of ORB-SLAM3~\cite{Campos2021ORBSLAM3}: the former estimates the current device pose, while the latter refines a covisibility-based local map. NIF-VIO uses visual--inertial consistency to adapt IMU fusion while maintaining a shared visual map across full and selective fusion modes. The following subsections detail the proposed method.

\subsection{Problem Formulation and Notation}
\label{sec:problem_formulation_notation}

We consider a device equipped with stereo cameras and an IMU. Our objective is to estimate device motion relative to the local rigid environment as inertial conditions change. Initialization is assumed to complete under inertial conditions, with carrier entry or exit occurring while the carrier is stationary and visual tracking remains continuous. Once onboard, the device observes a scene dominated by carrier-interior features. We focus on local odometry, leaving global mapping and loop closure across relatively moving environments outside the scope of this work.

In our formulation, $\underrightarrow{\mathcal F}_{\mathrm B}$, $\underrightarrow{\mathcal F}_{\mathrm E}$, and $\underrightarrow{\mathcal F}_{\mathrm W}$ denote the IMU body frame, local environment frame, and fixed inertial world frame, respectively. We use $\boldsymbol T^{\mathrm A}_{\mathrm B}\in\mathrm{SE}(3)$ to map coordinates from $\underrightarrow{\mathcal F}_{\mathrm B}$ to $\underrightarrow{\mathcal F}_{\mathrm A}$, with rotation $\boldsymbol R^{\mathrm A}_{\mathrm B}\in\mathrm{SO}(3)$ and translation $\boldsymbol p^{\mathrm A}_{\mathrm B}\in\mathbb R^3$. The system state is represented as
\begin{equation}
\mathcal X=
\left\{
\left\{
\boldsymbol T^{\mathrm E}_{\mathrm B,k},
\boldsymbol v_k^{\mathrm E},
\boldsymbol b_k^{\mathrm g},
\boldsymbol b_k^{\mathrm a}
\right\}_{k\in\mathcal K},
\left\{\boldsymbol\ell_j^{\mathrm E}\right\}_{j\in\mathcal L},
\boldsymbol g
\right\},
\label{eq:estimation_state}
\end{equation}
where $\mathcal K$ and $\mathcal L$ index local frames and landmarks. Velocity $\boldsymbol v$, landmarks $\boldsymbol\ell$, and gravity $\boldsymbol g$ are expressed in $\underrightarrow{\mathcal F}_{\mathrm E}$, while $\boldsymbol b^{\mathrm g}$ and $\boldsymbol b^{\mathrm a}$ denote gyroscope and accelerometer biases. The variables actively optimized depend on the fusion mode. Inside the carrier, the local frame is carrier-fixed, and the device-relative pose satisfies $\boldsymbol T^{\mathrm E}_{\mathrm B,k}=(\boldsymbol T^{\mathrm W}_{\mathrm E,k})^{-1}\boldsymbol T^{\mathrm W}_{\mathrm B,k}$, where $\boldsymbol T^{\mathrm W}_{\mathrm E,k}$ denotes the carrier pose. Visual observations of the interior constrain $\boldsymbol T^{\mathrm E}_{\mathrm B,k}$, whereas IMU measurements reflect both carrier motion and device-relative motion.

\subsection{Transition Management}
\label{sec:consistency_assessment_isolation}

As inertial conditions change, NIF-VIO adapts IMU fusion according to visual--inertial consistency. We form a joint residual $\boldsymbol e_n=[\boldsymbol e_{R,n}^{\top},\boldsymbol e_{p,n}^{\top}]^{\top}$ between image-based motion estimates and IMU preintegration to assess their agreement. The rotational component $\boldsymbol e_{R,n}$ is expressed using the $\mathrm{SO}(3)$ logarithm, while the translational component $\boldsymbol e_{p,n}$ is obtained by eliminating the unknown initial velocity from displacement relations across adjacent intervals. Since individual discrepancies may arise from measurement noise and estimation uncertainty, we compare two hypotheses: consistency within the expected uncertainty, and inconsistency involving additional motion disturbances. Both hypotheses are modeled using Student-$t$ distributions. The nominal density $p_0$ uses covariance obtained by uncertainty propagation, while $p_1$ is a mixture with additional disturbance covariance in rotation, translation, or both. Starting from $s_0=0$, we accumulate evidence as
\begin{equation}
\lambda_n=\log\frac{p_1(\boldsymbol e_n)}{p_0(\boldsymbol e_n)},
\qquad
s_n=\max(0,s_{n-1}+\lambda_n).
\label{eq:sequential_evidence}
\end{equation}
Evidence favoring inconsistency increases $s_n$, which otherwise decreases toward zero.

Reliable confirmation requires accumulated evidence, during which incompatible inertial constraints can affect local geometry. We therefore combine early constraint isolation to limit further error propagation with local reoptimization to correct accumulated geometric errors. Two thresholds on $s_n$ govern this process. The lower threshold temporarily excludes full IMU factors from tracking and local optimization while preserving visual constraints. Isolation is lifted only after subsequent observations validate visual--inertial consistency. The higher threshold confirms inconsistency and triggers removal of full IMU factors since the onset of evidence accumulation, followed by joint refinement of poses and landmarks within a bounded covisibility neighborhood. Starting from a visual-only estimate, we solve
\begin{equation}
\min_{\mathcal X_{\mathrm R},\,\boldsymbol u}
\left[
E_{\mathrm{vis}}(\mathcal X_{\mathrm R};\mathcal B)
+
E_{\mathrm{sel}}(\mathcal X_{\mathrm R},\boldsymbol u;\mathcal B)
\right].
\label{eq:local_repair}
\end{equation}
Here, $E_{\mathrm{vis}}$ is the robust reprojection cost over local poses and landmarks $\mathcal X_{\mathrm R}$, with outside observing keyframes held fixed as the boundary $\mathcal B$. The term $E_{\mathrm{sel}}$ introduces selective inertial constraints through auxiliary variables $\boldsymbol u$, jointly refining local geometry and carrier-motion estimates. These estimates initialize subsequent selective fusion, in which the same constraints continue to support local optimization. Their formulation is presented in the following subsection.

\subsection{Selective Inertial Constraints}
\label{sec:selective_inertial_constraints}

Under non-inertial conditions, IMU measurements still provide device-relative motion information that can complement visual constraints, but their use requires accounting for carrier-induced effects. During selective fusion, we represent these components through auxiliary parameters estimated within local optimization rather than recursively propagated system states. An unrestricted carrier model could absorb visual--inertial discrepancies without constraining device-relative motion. In practice, common carriers often exhibit local motion regularities. We incorporate this prior knowledge through temporal parameter sharing and soft motion constraints, limiting model flexibility to retain useful inertial information.

Considering gyroscope measurements first, mechanical and actuation constraints in common carriers often lead to rotation about a dominant axis with smoothly varying angular velocity. Over a local set of keyframes, we approximate this variation by a linear model. The carrier angular velocity relative to the inertial frame, expressed in $\underrightarrow{\mathcal F}_{\mathrm E}$, is modeled as
\begin{equation}
\boldsymbol\Omega^{\mathrm E}(t)=\boldsymbol\Omega_c+\boldsymbol\alpha(t-t_c).
\label{eq:carrier_angular_model}
\end{equation}
Here, $\boldsymbol\Omega_c$ and $\boldsymbol\alpha$ denote the angular velocity at the interval midpoint $t_c$ and its rate of change, respectively, both expressed in $\underrightarrow{\mathcal F}_{\mathrm E}$. An unknown unit vector $\boldsymbol n$ defines the dominant rotation axis, and $\boldsymbol P_\perp=\boldsymbol I-\boldsymbol n\boldsymbol n^\top$ extracts off-axis components. We impose the soft regularizer
\begin{equation}
\Phi_\omega=\frac{\|\boldsymbol\alpha\|^2}{\sigma_\alpha^2}
+\frac{\|\boldsymbol P_\perp\boldsymbol\Omega_c\|^2}{\sigma_{\omega,\perp}^2}
+\frac{\|\boldsymbol P_\perp\boldsymbol\alpha\|^2}{\sigma_{\alpha,\perp}^2}.
\label{eq:carrier_motion_regularizer}
\end{equation}
The first term penalizes angular-velocity changes, while the other two penalize off-axis components of $\boldsymbol\Omega_c$ and $\boldsymbol\alpha$. The fixed scale parameters $\sigma_\alpha$, $\sigma_{\omega,\perp}$, and $\sigma_{\alpha,\perp}$ control the respective regularization strengths. When rotational excitation is insufficient, the dominant-axis direction is held fixed.

For adjacent keyframes $i$ and $j$ within the modeled interval, integrating the carrier angular-velocity model yields their relative carrier rotation $\boldsymbol U_{ij}(\boldsymbol\Omega_c,\boldsymbol\alpha)$. Writing $\boldsymbol R_i=\boldsymbol R^{\mathrm E}_{\mathrm B,i}$, the pose composition predicts the relative device rotation with respect to the inertial frame as $\boldsymbol R_i^\top\boldsymbol U_{ij}\boldsymbol R_j$. Comparing this prediction with IMU preintegration defines the rotation residual
\begin{equation}
\boldsymbol r^R_{ij}=\operatorname{Log}\!\left[
\Delta\boldsymbol R_{ij}(\boldsymbol b^{\mathrm g})^\top
\boldsymbol R_i^\top\boldsymbol U_{ij}(\boldsymbol\Omega_c,\boldsymbol\alpha)\boldsymbol R_j
\right].
\label{eq:selective_rotation_residual}
\end{equation}
During non-inertial operation, the bias $\boldsymbol b^{\mathrm g}$ is held fixed at its most recent estimate supported by visual--inertial consistency to prevent unmodeled carrier motion from being absorbed into the bias estimate. We stack these residuals into $\boldsymbol r$, with covariance $\boldsymbol S$ accounting for preintegration noise and shared bias uncertainty. Combining the inertial residuals, robust visual cost, and carrier-motion regularizer leads to the joint optimization objective:
\begin{equation}
\min_{\mathcal X_{\mathrm{loc}},\,\boldsymbol u}
E_{\mathrm{vis}}(\mathcal X_{\mathrm{loc}})
+\boldsymbol r^\top\boldsymbol S^{-1}\boldsymbol r+\Phi,
\label{eq:selective_objective}
\end{equation}
where $\mathcal X_{\mathrm{loc}}$ contains local poses and landmarks, and $\boldsymbol u$ collects auxiliary parameters. For gyroscope retention, $\boldsymbol u=(\boldsymbol\Omega_c,\boldsymbol\alpha,\boldsymbol n)$, with $\|\boldsymbol n\|=1$ and $\Phi=\Phi_\omega$.

Building on the carrier-rotation model, we further incorporate accelerometer measurements to constrain relative translation. Let $\boldsymbol Q_i$ map vectors from $\underrightarrow{\mathcal F}_{\mathrm E}$ to a local nonrotating frame $\underrightarrow{\mathcal F}_{\mathrm N}$. Its axes initially align with the carrier axes, so $\boldsymbol Q_0=\boldsymbol I$, and their relation is updated as $\boldsymbol Q_j=\boldsymbol Q_i\boldsymbol U_{ij}$. For device position $\boldsymbol p_i=\boldsymbol p^{\mathrm E}_{\mathrm B,i}$ and a reference point $\boldsymbol p_c$ fixed in the same frame, we express the relative position and body orientation in the nonrotating frame as $\boldsymbol y_i=\boldsymbol Q_i(\boldsymbol p_i-\boldsymbol p_c)$ and $\boldsymbol D_i=\boldsymbol Q_i\boldsymbol R_i$. The corresponding continuous-time dynamics, omitting measurement noise, are
\begin{equation}
\ddot{\boldsymbol y}
=\boldsymbol D(\boldsymbol f_m-\boldsymbol b^{\mathrm a})-\boldsymbol h,
\qquad
\boldsymbol h\equiv\boldsymbol h^{\mathrm N}=\boldsymbol a_c^{\mathrm N}-\boldsymbol g^{\mathrm N},
\label{eq:transformed_acceleration}
\end{equation}
where $\boldsymbol f_m$ is measured specific force, and $\boldsymbol a_c^{\mathrm N}$ and $\boldsymbol g^{\mathrm N}$ denote the reference-point acceleration and gravity expressed in $\underrightarrow{\mathcal F}_{\mathrm N}$. Only their combined effect $\boldsymbol h$ is estimated.

Over the local interval, we approximate the combined effect of carrier acceleration and gravity by a temporally shared linear model, $\boldsymbol h(t)=\boldsymbol h_c+\boldsymbol j(t-t_c)$, where $\boldsymbol h_c$ and $\boldsymbol j$ are its midpoint value and rate of change in $\underrightarrow{\mathcal F}_{\mathrm N}$. The soft regularizer $\Phi_a=\|\boldsymbol j\|^2/\sigma_j^2$ favors gradual variation, with $\sigma_j$ controlling its strength. Defining the auxiliary velocity $\boldsymbol w=\dot{\boldsymbol y}$ and interval duration $T_{ij}=t_j-t_i$, we integrate the transformed dynamics to obtain
\begin{equation}
\begin{aligned}
\boldsymbol r^v_{ij}
&=\boldsymbol D_i^\top
\left(\boldsymbol w_j-\boldsymbol w_i+\boldsymbol H_{ij}\right)
-\Delta\boldsymbol v_{ij},\\
\boldsymbol r^p_{ij}
&=\boldsymbol D_i^\top
\left(\boldsymbol y_j-\boldsymbol y_i-\boldsymbol w_iT_{ij}
+\boldsymbol K_{ij}\right)
-\Delta\boldsymbol p_{ij},
\end{aligned}
\label{eq:selective_translation_residual}
\end{equation}
where $\boldsymbol H_{ij}=\int_{t_i}^{t_j}\boldsymbol h(t)\,dt$ and $\boldsymbol K_{ij}=\int_{t_i}^{t_j}(t_j-t)\boldsymbol h(t)\,dt$. The IMU preintegrals $\Delta\boldsymbol v_{ij}$ and $\Delta\boldsymbol p_{ij}$ use fixed gyroscope and accelerometer biases. These velocity and position residuals enter the joint objective alongside the rotation residuals, with their correlations captured by $\boldsymbol S$ and $\Phi=\Phi_\omega+\Phi_a$. Accordingly, $\boldsymbol u$ is extended to include $\boldsymbol h_c$, $\boldsymbol j$, and $\{\boldsymbol w_i\}$.

The joint formulation uses auxiliary variables to account for carrier-induced motion. To examine their effect on relative-motion constraints, we locally eliminate these variables. At fixed linearization and weights, the whitened inertial and regularization residuals are $\bar{\boldsymbol r}+\boldsymbol A\delta\boldsymbol x+\boldsymbol B\delta\boldsymbol u$, where $\delta\boldsymbol x$ and $\delta\boldsymbol u$ are local tangent-space increments of the active variables, with corresponding Jacobians $\boldsymbol A$ and $\boldsymbol B$. With $\boldsymbol N$ an orthonormal basis for $\operatorname{null}(\boldsymbol B^\top)$,
\begin{equation}
\min_{\delta\boldsymbol u}
\left\|\bar{\boldsymbol r}+\boldsymbol A\delta\boldsymbol x
+\boldsymbol B\delta\boldsymbol u\right\|^2
=
\left\|\boldsymbol N^\top
\left(\bar{\boldsymbol r}+\boldsymbol A\delta\boldsymbol x\right)\right\|^2.
\label{eq:retained_inertial_information}
\end{equation}
After elimination, only residual components that cannot be absorbed by the auxiliary variables contribute to the reduced cost. Temporal sharing and soft priors limit this absorption, leaving constraints along relative-motion directions for which $\boldsymbol N^\top\boldsymbol A\delta\boldsymbol x\neq\boldsymbol 0$. This characterizes the locally retained information without implying full observability, and unconstrained directions still rely on visual support.

\begin{figure*}[t]
\centering
\includegraphics[width=0.9\textwidth]{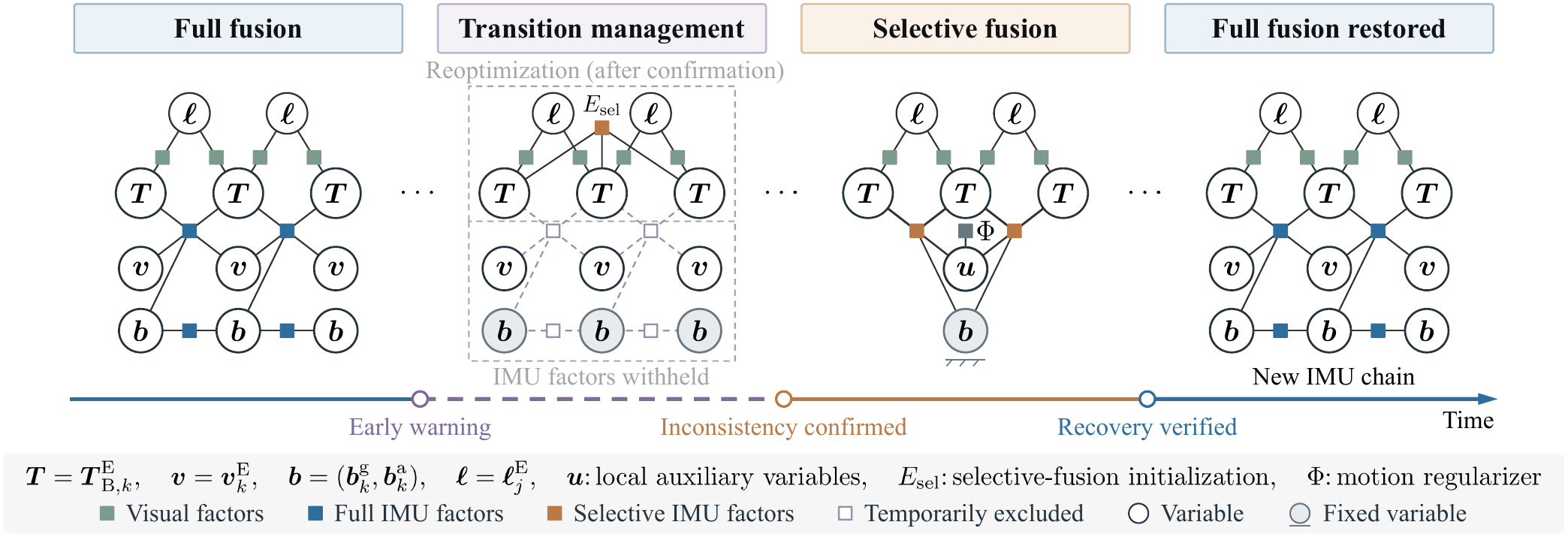}
\caption{Evolution of the local factor graph through a representative sequence of fusion stages. Each configuration shows the active constraints and optimized variables, illustrating how visual--inertial fusion adapts to changing inertial conditions.}
\label{fig:fusion_factor_graph}
\end{figure*}

\subsection{Fusion Recovery and Factor-Graph Evolution}
\label{sec:full_fusion_restoration}

When inertial conditions return, we seek to restore full use of IMU information. This requires velocity and gravity estimates consistent with the full inertial model, which the auxiliary quantities of selective fusion do not directly provide. We estimate the initial velocity $\boldsymbol v_0^{\mathrm E}$ and gravity $\boldsymbol g$ in $\underrightarrow{\mathcal F}_{\mathrm E}$ by fitting IMU preintegration to visual poses over a sliding window, with biases fixed and gravity magnitude constrained. Since carrier acceleration can be absorbed into the fitted gravity, a small fitting residual alone does not establish recovery. The estimated state is therefore propagated using new IMU measurements and checked against subsequent visual observations. We accept recovery based on sustained predictive agreement and renewed visual--inertial consistency, using the validated state to initialize a full IMU chain and restore IMU-based motion prediction and visual--inertial tracking.

With full fusion restored, the framework completes a cycle of adapting inertial constraints to changing inertial conditions. Fig.~\ref{fig:fusion_factor_graph} summarizes this process through a representative evolution of the local factor graph. Visual factors remain active throughout, providing a common geometric basis across stages. As evidence of inconsistency accumulates, early isolation temporarily withholds full IMU factors to limit further error propagation. Confirmation then triggers local reoptimization to correct accumulated geometric errors. Subsequent selective fusion preserves useful inertial support for relative-motion estimation through the carrier-motion model. Once recovery is verified, visual--inertial fusion resumes with a full IMU constraint chain.

\section{EXPERIMENTS}
\label{sec:experiments}

This section evaluates the overall performance of NIF-VIO and the effectiveness of its key components. We assess localization accuracy and availability through comparative experiments, followed by ablation studies, runtime analysis, and an augmented reality (AR) demonstration under non-inertial conditions.

\begin{table}[t]
\centering
\caption{Localization Accuracy of VIO Methods on the EuRoC and TUM-VI Datasets}
\label{tab:public_localization}
\footnotesize
\renewcommand{\arraystretch}{1.05}
\setlength{\aboverulesep}{0.3ex}
\setlength{\belowrulesep}{0.45ex}
\setlength{\tabcolsep}{2.5pt}
\begin{tabular*}{\columnwidth}{@{\extracolsep{\fill}}l*{6}{c}@{}}
\toprule[1pt]
\multicolumn{1}{@{}c}{\multirow{3}{*}[-2pt]{Method}} & \multicolumn{2}{c}{Machine Hall} & \multicolumn{2}{c}{Vicon Room} & \multicolumn{2}{c}{TUM-VI Room} \\
\cmidrule(l{0.25em}r{0.25em}){2-3}\cmidrule(l{0.25em}r{0.25em}){4-5}\cmidrule(l{0.25em}){6-7}
 & ATE & ARE & ATE & ARE & ATE & ARE \\
 & (m) $\downarrow$ & ($^\circ$) $\downarrow$ & (m) $\downarrow$ & ($^\circ$) $\downarrow$ & (m) $\downarrow$ & ($^\circ$) $\downarrow$ \\
\midrule
VINS-Fusion & 0.282 & 1.945 & 0.148 & 3.232 & 0.071 & 1.171 \\
OpenVINS & 0.138 & 1.674 & 0.065 & 1.656 & 0.074 & 1.984 \\
Basalt & 0.088 & \textbf{1.074} & 0.076 & \textbf{1.317} & 0.066 & 1.989 \\
OKVIS2 & 0.087 & 1.261 & \textbf{0.043} & 1.403 & 0.044 & 1.139 \\
SchurVINS & 0.105 & 1.585 & 0.070 & 2.137 & 0.073 & 2.026 \\
$\sqrt{\mathrm{VINS}}$ & 0.100 & 1.555 & 0.054 & 1.645 & 0.072 & 1.910 \\
ORB-SLAM3 & \underline{0.075} & \underline{1.228} & 0.058 & 1.668 & \underline{0.022} & \textbf{0.525} \\
\midrule
NIF-VIO (Ours) & \textbf{0.065} & 1.239 & \underline{0.049} & \underline{1.392} & \textbf{0.019} & \underline{0.550} \\
\bottomrule[1pt]
\end{tabular*}
\par\smallskip
\begin{minipage}{\columnwidth}
\footnotesize
Results are averaged over five runs per sequence, then equally across sequences within each scene group.
\textbf{Bold} and \underline{underlined} values denote the best and second-best results in each column.
We reconstruct the EuRoC V1\_01\_easy reference using the method in~\cite{Shu2025HPGT} to address its orientation inaccuracies~\cite{Geneva2020OpenVINS}.
\end{minipage}
\end{table}

\subsection{Dataset}
\label{sec:carrier_relative_dataset}

To evaluate carrier-relative localization under changing inertial conditions, we collect a dedicated dataset covering car, subway, and elevator scenarios. Each scenario includes five sequences, all of which begin under inertial conditions, with the carrier interior becoming the dominant visual reference before motion onset. Carrier acceleration or rotation then introduces non-inertial conditions, followed by a return to inertial conditions when the carrier eventually stops. We record the dataset using an Intel RealSense T265, providing $848\times800$ stereo fisheye images at 30~Hz and IMU measurements at 200~Hz. Camera--IMU calibration is performed offline, and IMU noise parameters are determined through Allan variance analysis.

Obtaining high-accuracy reference trajectories inside moving carriers using optical motion capture (MoCap) or similar outside-in tracking systems poses practical deployment challenges. We therefore use known-size ArUco markers to assist reference trajectory estimation. The markers are placed in low-texture regions to minimize occlusion of natural features, and their image regions are masked during evaluation. We obtain reference trajectories in the local environment frame through offline global bundle adjustment (BA), using marker and natural-feature observations from the onboard stereo cameras. The resulting poses are screened for visual support and geometric consistency, with the retained reference poses covering an average of 98.74\% of the total input duration per sequence. We validate the same pipeline against MoCap under observation conditions similar to those of our dataset, obtaining absolute translation and rotation errors (ATE and ARE) of 2.1~mm and 0.18$^\circ$, and relative translation and rotation errors (RTE and RRE) of 0.9~mm and 0.07$^\circ$ over 0.5~s intervals. All errors are reported as root mean square errors (RMSEs).

\subsection{Localization Performance}
\label{sec:localization_performance}

\begin{table*}[t]
\centering
\caption{Localization Accuracy and Correct Localization Rate of VO and VIO Methods on Our Dataset}
\label{tab:self_localization}
\renewcommand{\arraystretch}{0.95}
\setlength{\aboverulesep}{0.3ex}
\setlength{\belowrulesep}{0.45ex}
\setlength{\tabcolsep}{1.5pt}
\begin{tabular*}{0.97\textwidth}{@{\extracolsep{\fill}}cl*{9}{c}@{}}
\toprule[1pt]
\multirow{2}{*}[-2pt]{Mode} & \multicolumn{1}{c}{\multirow{2}{*}[-2pt]{Method}} & \multicolumn{3}{c}{Car (5 seq.)} & \multicolumn{3}{c}{Subway (5 seq.)} & \multicolumn{3}{c}{Elevator (5 seq.)} \\
\cmidrule(l{0.25em}r{0.25em}){3-5}\cmidrule(l{0.25em}r{0.25em}){6-8}\cmidrule(l{0.25em}){9-11}
 & & C-ATE (m) $\downarrow$ & C-ARE ($^\circ$) $\downarrow$ & CR (\%) $\uparrow$ & C-ATE (m) $\downarrow$ & C-ARE ($^\circ$) $\downarrow$ & CR (\%) $\uparrow$ & C-ATE (m) $\downarrow$ & C-ARE ($^\circ$) $\downarrow$ & CR (\%) $\uparrow$ \\
\midrule
\multirow{4}{*}{VO} & VINS-Fusion & 0.038 & 3.413 & 64.83 & 0.081 & 2.627 & 83.09 & 0.125 & 2.786 & 77.85 \\
 & Basalt & 0.042 & 2.198 & 57.45 & 0.085 & 2.072 & 56.46 & 0.142 & 4.174 & 70.21 \\
 & OKVIS2 & 0.020 & 0.861 & 68.95 & \underline{0.031} & \textbf{0.670} & 72.31 & 0.089 & 1.685 & 84.78 \\
 & ORB-SLAM3 & 0.019 & 1.208 & 93.43 & 0.034 & 1.008 & 94.83 & 0.094 & 1.495 & 89.42 \\
\midrule
\multirow{7}{*}{VIO} & VINS-Fusion & 0.037 & 1.279 & 38.72 & 0.131 & 1.478 & 20.76 & 0.111 & 1.944 & 35.54 \\
 & OpenVINS & 0.045 & 0.936 & 22.19 & 0.061 & 0.902 & 22.17 & 0.097 & 1.230 & 37.61 \\
 & Basalt & 0.053 & 2.024 & 51.92 & 0.099 & 2.693 & 87.55 & 0.108 & 1.514 & 93.98 \\
 & OKVIS2 & 0.026 & 0.950 & 27.38 & 0.060 & 0.847 & 25.60 & 0.086 & 1.286 & 32.81 \\
 & SchurVINS & 0.069 & 1.193 & 26.87 & 0.190 & 1.887 & 40.05 & 0.116 & 1.156 & 43.30 \\
 & $\sqrt{\mathrm{VINS}}$ & 0.052 & 0.848 & 22.32 & 0.127 & 0.856 & 17.91 & 0.132 & 1.328 & 31.66 \\
 & ORB-SLAM3 & 0.066 & 0.851 & 24.00 & 0.124 & 1.301 & 35.79 & 0.100 & 1.236 & 76.61 \\
\midrule
\multirow{3}{*}{Adaptive} & NIF-VIO (w/o TM) & 0.039 & 0.917 & 72.19 & 0.068 & 0.905 & 78.31 & 0.092 & \underline{1.031} & \underline{98.09} \\
 & NIF-VIO (w/o SF) & \textbf{0.010} & \underline{0.708} & \underline{96.45} & 0.033 & 0.792 & \underline{95.21} & \underline{0.083} & 1.337 & 94.97 \\
 & NIF-VIO (Ours) & \underline{0.012} & \textbf{0.671} & \textbf{98.49} & \textbf{0.025} & \underline{0.677} & \textbf{98.56} & \textbf{0.070} & \textbf{0.943} & \textbf{98.19} \\
\bottomrule[1pt]
\end{tabular*}
\par\smallskip
\begin{minipage}{0.97\textwidth}
\footnotesize
Results are averaged over five runs per sequence, then equally across sequences within each scene group.
\textbf{Bold} and \underline{underlined} values denote the best and second-best results in each column.
C-ATE and C-ARE are computed over correct outputs only.
\end{minipage}
\end{table*}

We compare NIF-VIO with seven representative odometry systems: VINS-Fusion~\cite{Qin2019VINSFusion}, OpenVINS~\cite{Geneva2020OpenVINS}, Basalt~\cite{Usenko2020Basalt}, ORB-SLAM3~\cite{Campos2021ORBSLAM3}, OKVIS2~\cite{Leutenegger2022OKVIS2}, SchurVINS~\cite{Fan2024SchurVINS}, and $\sqrt{\mathrm{VINS}}$~\cite{Peng2025SqrtVINS}. All methods use stereo inputs with fixed calibration. To focus on online odometry, we disable loop closure and map merging and evaluate the original online pose outputs. Experiments run on an AMD Ryzen 9 7945HX computer with 32~GB RAM and Ubuntu 20.04. In our implementation, early isolation and inconsistency confirmation in Section~\ref{sec:consistency_assessment_isolation} are triggered when $s_n$ reaches 2 and 6, respectively. The carrier model in Section~\ref{sec:selective_inertial_constraints} shares parameters over at most 2~s, with regularization scales $\sigma_\alpha=0.10\,\mathrm{rad/s^2}$, $\sigma_{\omega,\perp}=0.03\,\mathrm{rad/s}$, $\sigma_{\alpha,\perp}=0.03\,\mathrm{rad/s^2}$, and $\sigma_j=1.0\,\mathrm{m/s^3}$.

\begin{figure*}[t]
\centering
\includegraphics[width=0.92\textwidth]{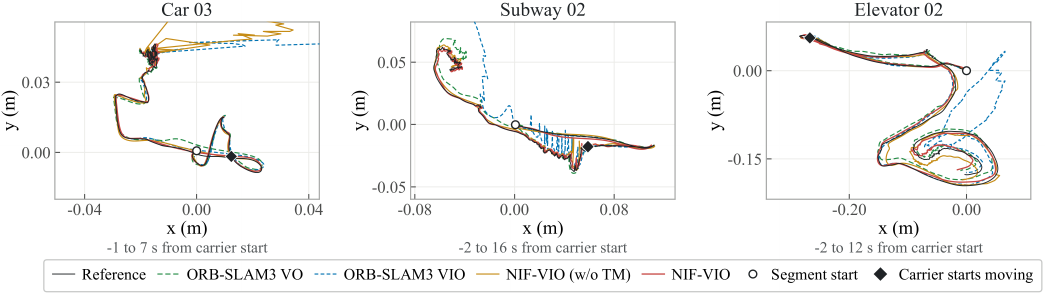}
\caption{Planar views of carrier-relative trajectory segments around carrier motion onset in three representative sequences. Selected methods are compared against the reference trajectories to illustrate tracking behavior.}
\label{fig:transition_trajectories}
\end{figure*}

We first assess localization accuracy in conventional inertial scenarios using the Machine Hall and Vicon Room sequences from the EuRoC dataset~\cite{Burri2016EuRoC} and the room sequences from the TUM-VI dataset~\cite{Schubert2018TUMVI}. ATE and ARE are computed after aligning the estimated trajectories with the ground truth using an $\mathrm{SE}(3)$ transformation. For compact presentation, results are averaged over five runs per sequence and then equally across sequences within each scene group, as reported in Table~\ref{tab:public_localization}. NIF-VIO achieves the best or second-best translation accuracy across all three scene groups, together with competitive rotation accuracy. Across these runs, no false confirmations of visual--inertial inconsistency were observed. These results show that NIF-VIO maintains accurate localization in conventional inertial scenarios.

We next evaluate localization across changing inertial conditions on our dataset. The comparison methods are also tested in their available stereo VO modes. We introduce two variants of NIF-VIO to assess its key components: NIF-VIO (w/o TM) replaces transition management with direct switching upon confirmed inconsistency, while NIF-VIO (w/o SF) disables selective inertial fusion and uses only visual constraints for pose estimation under non-inertial conditions. To account for possible trajectory divergence and resets, we select the estimated trajectory with the longest output time coverage in a single coordinate frame and align it to the reference using a single $\mathrm{SE}(3)$ transformation. This transformation is iteratively refined through residual-based rejection of outlying pose correspondences, while all outputs are retained for subsequent evaluation. After alignment, outputs with translation and rotation errors within 0.20~m and $5^\circ$, respectively, are classified as correct. The correct localization rate (CR) is then defined as the ratio of correctly localized duration to the total duration covered by reference poses. Conditional ATE (C-ATE) and ARE (C-ARE) are computed over the correct outputs of each method and interpreted together with CR.

As shown in Table~\ref{tab:self_localization}, visual--inertial conflicts under non-inertial conditions lead to substantially reduced CR for most conventional VIO methods. VO avoids these conflicts but cannot benefit from complementary inertial information. NIF-VIO maintains CR above 98\% across all three scenarios with low conditional errors. Comparisons with the w/o TM and w/o SF variants indicate that transition management and selective inertial fusion contribute to overall improvements in localization accuracy and availability. Differences between methods are particularly apparent around carrier motion onset, as illustrated by the trajectory segments from three representative sequences in Fig.~\ref{fig:transition_trajectories}. ORB-SLAM3 in VIO mode shows pronounced deviations from the reference, whereas its VO mode remains stable but retains some estimation error. The w/o TM variant shows a noticeable deviation in the car example. In comparison, the full method follows the reference more closely across the three examples.

\subsection{Ablation Studies}
\label{sec:ablation_studies}

\begin{figure}[t]
\centering
\includegraphics[width=0.98\columnwidth]{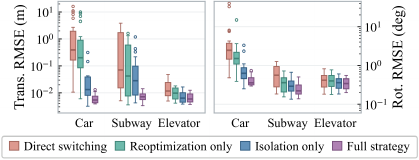}
\caption{Ablation of the transition-handling strategy. Translation and rotation RMSE distributions summarize pose estimation errors around the transition to non-inertial conditions for four configurations in each scenario.}
\label{fig:transition_strategy_ablation}
\end{figure}

\begin{figure}[t]
\centering
\includegraphics[width=\columnwidth]{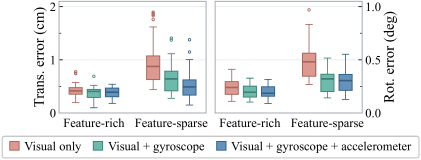}
\caption{Ablation of IMU information retention. Endpoint-relative translation and rotation error distributions are shown for three configurations in feature-rich and feature-sparse windows during non-inertial operation.}
\label{fig:imu_retention_ablation}
\end{figure}

\begin{figure}[t]
\centering
\includegraphics[width=\columnwidth]{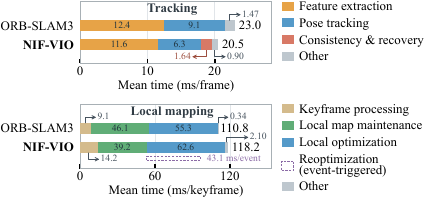}
\caption{Mean processing times of NIF-VIO and ORB-SLAM3 in VIO mode. Tracking and local mapping are reported separately, with reoptimization time measured per event.}
\label{fig:runtime_analysis}
\end{figure}

\begin{figure}[t]
\centering
\includegraphics[width=\columnwidth]{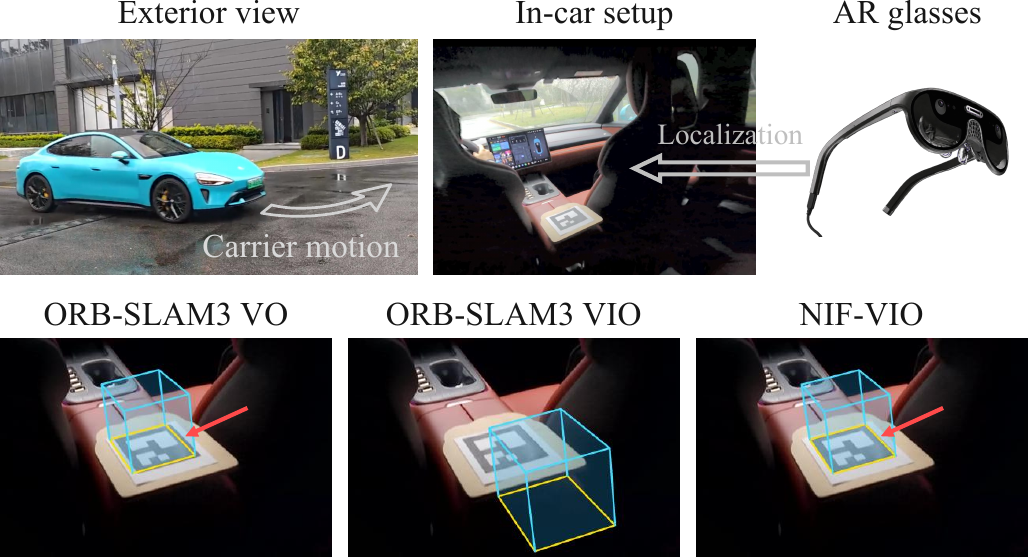}
\caption{AR demonstration inside a car. The upper row shows the experimental setup, and the lower row compares virtual cube renderings from different localization methods at the same instant after the car starts moving. Red arrows highlight differences in virtual cube alignment between NIF-VIO and ORB-SLAM3 operating in VO mode.}
\label{fig:ar_demo}
\end{figure}

We further analyze the key components through two ablation studies. For transition management (Section~\ref{sec:consistency_assessment_isolation}), we compare four configurations: direct switching, local reoptimization only, early IMU factor isolation only, and their combination, with five runs per sequence and configuration. To focus on the transition, we evaluate pose errors within a common window from 1~s before the initial inconsistency warning to 3~s after confirmation and report translation and rotation RMSEs. The alignment transformation is estimated from the pre-warning segment. As shown in Fig.~\ref{fig:transition_strategy_ablation}, both isolation and reoptimization reduce transition errors in the car and subway sequences, while their combination achieves the lowest median errors across all three scenarios. The differences between configurations are smaller in the elevator sequences, consistent with their relatively mild motion transitions.

The second ablation examines the benefit of retained inertial information during non-inertial operation. Following Section~\ref{sec:selective_inertial_constraints}, we compare visual-only estimation, retention of gyroscope measurements, and the further inclusion of accelerometer constraints. To assess their effects under different visual conditions, we sample ten non-overlapping 0.5~s windows from each sequence and rank them by the median per-frame image coverage of tracked inliers from the visual-only configuration. The two highest- and lowest-ranked windows form the feature-rich and feature-sparse groups, respectively. Fig.~\ref{fig:imu_retention_ablation} compares endpoint-relative translation and rotation errors for the three configurations in each group. Retaining gyroscope measurements reduces translation and rotation errors in feature-sparse windows. Adding accelerometer constraints further reduces translation error, with little change in rotation error. The three configurations yield similar errors when visual features are abundant. These results suggest that retained IMU information helps compensate for limited visual constraints, improving pose estimation.

\subsection{Runtime Analysis}
\label{sec:runtime_analysis}

The computational cost of NIF-VIO is evaluated by averaging runtime measurements across sequences in our dataset. ORB-SLAM3 in VIO mode serves as a reference due to its similar tracking and local mapping architecture. Fig.~\ref{fig:runtime_analysis} reports processing times separately for the asynchronous tracking and local mapping threads. NIF-VIO spends less time on conventional pose tracking and local map maintenance. Consistency assessment and fusion recovery add an average overhead of 1.64~ms per frame. Selective inertial fusion increases local optimization time through additional variables and constraints, while transition-triggered reoptimization incurs a separate cost upon confirmed inconsistency. Overall, the mean tracking time remains below the camera frame interval, with a moderate increase in ordinary local mapping time.

\subsection{AR Demonstration}
\label{sec:ar_demo}

To illustrate a practical application of NIF-VIO, we present an AR example inside a moving car using data captured by our AR glasses. We compare NIF-VIO with ORB-SLAM3 in VO and VIO modes, using their estimated poses to overlay a virtual cube on color images from the onboard video see-through camera. While the car is stationary, an ArUco marker initializes the same cube placement for all methods. Marker regions are then masked during localization, and subsequent rendering relies solely on the estimated poses. Fig.~\ref{fig:ar_demo} compares the AR results at the same instant after the car starts moving. ORB-SLAM3 in VIO mode shows a pronounced cube displacement due to carrier-induced visual--inertial conflicts. Its VO mode avoids these conflicts but exhibits an anchoring error under low-light, low-texture conditions inside the car. NIF-VIO better preserves the intended placement, illustrating its ability to maintain accurate localization relative to the carrier under non-inertial conditions.

\section{CONCLUSION}
\label{sec:conclusion}

This paper presented NIF-VIO for continuous state estimation across inertial and non-inertial conditions. The framework adapts the use of inertial measurements according to visual--inertial consistency. By coordinating early constraint isolation and local reoptimization, it addresses both error propagation and the resulting distortion of local geometry as inertial conditions change. Under non-inertial conditions, temporal parameter sharing and soft motion priors constrain the carrier-motion model, allowing inertial information to continue supporting relative-motion estimation. Experiments demonstrate competitive performance under inertial conditions and improved localization accuracy and availability in non-inertial scenarios. Future work will extend NIF-VIO beyond scenes dominated by the carrier interior, using external visual observations to better constrain carrier motion and strengthen inertial fusion for relative localization.

\bibliographystyle{bib/IEEEtran}
\bibliography{bib/controls,bib/references}

\end{document}